%% file: example_paper.tex
\documentclass{article}

\usepackage{microtype}
\usepackage{graphicx}
\usepackage{subcaption}
\usepackage{booktabs} 
\usepackage{multirow}
\usepackage[table,xcdraw]{xcolor}
\usepackage{hyperref}

\usepackage[preprint]{icml2026}

\usepackage{amsmath}
\usepackage{amssymb}
\usepackage{mathtools}
\usepackage{amsthm}

\usepackage{tikz}
\usepackage{subcaption}

\usepackage{pgfplots}

\usepackage{caption,subcaption}

\usepackage[dvipsnames]{xcolor}
\usepackage[many]{tcolorbox}
\usepackage{enumitem}

\usepackage[capitalize,noabbrev]{cleveref}

\theoremstyle{plain}

\theoremstyle{definition}

\theoremstyle{remark}

\usepackage[textsize=tiny]{todonotes}

\icmltitlerunning{Segmentation before Answering: Pixel Grounding for MLLM Visual Reasoning}

\begin{document}

\twocolumn[
  \icmltitle{Segmentation before Answering: Pixel Grounding for MLLM Visual Reasoning}



  \icmlsetsymbol{equal}{*}

\begin{icmlauthorlist}
\icmlauthor{Yake Wei}{yyy,comp}
\icmlauthor{Yuan Wang}{ccc}
\icmlauthor{Fengyun Rao}{ddd}
\icmlauthor{Jing LYU}{ddd}
\icmlauthor{Di Hu}{yyy,comp}
\end{icmlauthorlist}

\icmlaffiliation{yyy}{Gaoling School of Artificial Intelligence, Renmin University of China}
\icmlaffiliation{comp}{Laboratory of Research on Large Models and Intelligent Governance}
\icmlaffiliation{ccc}{Tsinghua University}
\icmlaffiliation{ddd}{WeChat Vision, Tencent}

\icmlcorrespondingauthor{Di Hu}{dihu@ruc.edu.cn}

  \icmlkeywords{Machine Learning, ICML}

  \vskip 0.3in
]



\printAffiliationsAndNotice{}  

\input{version3/abstract}

\input{version3/introduction}

\input{version3/relatedwork}

\input{version3/method}

\input{version3/experiments}

\input{version3/conclusion}

\bibliography{example_paper}
\bibliographystyle{icml2026}




\end{document}

%% file: version3/abstract.tex
\begin{abstract}
Recent advancements in Multimodal Large Language Models (MLLMs) have evolved from static perception to interleaved visual-language reasoning, often referred to as ``thinking with images''. A basic operation in this reasoning process is to zoom in on regions of interest (often represented with bounding boxes) to acquire finer visual details. In this paper, we propose \textbf{Seg}mentation before \textbf{Answer}ing (SegAnswer), which shifts the unit of zoom-in from the popular bounding box to pixel-level segmentation mask. By employing fine-grained masks to isolate the target area from cluttered environments, segmented visual input yields a more precise region of interest, effectively filtering out redundant background and interfering objects. Furthermore, the discrete patches of segmented visual input align more seamlessly with how MLLMs structure visual tokens via positional embeddings. In experiments, we evaluate SegAnswer across diverse benchmarks, including high-resolution perception, general perception, and hallucination. It achieves consistent improvements and also exhibits considerable performance on segmentation tasks, validating its capability for reliable pixel grounding.
\end{abstract}

%% file: version3/introduction.tex
\section{Introduction}

Recent advancements in Multimodal Large Language Models (MLLMs) have demonstrated remarkable progress across a broad range of vision-language tasks~\cite{li2024llava,bai2025qwen2}, covering from fundamental image captioning to complex visual question answering. Moving beyond static visual perception where the visual inputs of MLLMs are fixed, immutable contexts processed in a single forward pass, explorations about interleaved visual-language reasoning have raised wide attention recently~\cite{wang2025pixel,zheng2025deepeyes}. These methods are often referred to as ``thinking with images''. In these methods, models often actively manipulate visual inputs to capture and re-examine regions of interest, thereby acquiring finer details beyond the original input image to generate the final answer.

Among the various operations employed when thinking with images, a basic operation in this reasoning process is to zoom in on regions of interest~\cite{wang2025pixel,zheng2025deepeyes}. Concretely, when executing a zoom-in operation, the model would typically generate bounding boxes (BBox) that delineate the area. Subsequently, this localized area is cropped from the original image, serving as a new visual input for the model's next step of reasoning. Despite its widespread adoption, this kind of BBox-based zoom-in operation encounters significant and often overlooked challenges in the complex practical applications.

\input{table_figure/fig_teaser}

As illustrated in~\autoref{fig:teasera}, a primary limitation arises from the inherent irregularity of target object shapes. The rigid rectangular prior of bounding boxes fails to capture the diverse geometric characteristics of natural objects, which are often irregular. Consequently, rectangular boxes inevitably encompass substantial redundant background, resulting in inefficient token consumption. Furthermore, in scenarios involving overlapping objects, simple bounding boxes often fail to disentangle the target from adjacent entities. This inability to precisely isolate the region of interest would lead to semantic interference, where the model struggles to distinguish the intended object from its surroundings. Fundamentally, these issues stem from the coarse granularity of bounding boxes, which lack the capacity to provide a precise observation of the region of interest.

To mitigate these limitations of the current BBox-based zoom-in operation, we shift from rectangular bounding boxes to pixel-level segmentation masks. As shown in~\autoref{fig:teaserb}, this segmentation-based pixel grounding requires the model to generate a fine-grained and accurate segmentation mask of the region of interest, instead of coarse and rectangular bounding boxes. By leveraging these segmented visual inputs, it effectively filters out redundant background and extraneous visual signals, ensuring that only the regions of interest are preserved for subsequent reasoning. Furthermore, for the newly introduced segmented visual input, we keep the position index of each segmented image patch as it is in the original image, thereby sparse segmented image patches can effectively reflect spatial relations. This also aligns more seamlessly with the nature that MLLMs structure visual tokens via positional embedding. Then, these cleaner, unambiguous visual inputs are further fed into models to provide finer details, enhancing visual reasoning.

Based on the above pixel grounding, we propose the \textbf{Seg}mentation before \textbf{Answer}ing (SegAnswer) method in this paper, which employs the segmentation during the visual reasoning process. Concretely, the training of SegAnswer has three stages. We first conduct pixel grounding to equip the model with semantic segmentation ability. Then, a multimodal interleaved supervised fine-tuning is conducted to instruct the model to perform pixel grounding as an intermediate conversation step, utilizing the segmented images for generating the final answer. Finally, reasoning with pixel grounding is carried out to enhance the MLLM visual reasoning by reinforcement learning.

In experiments, we first evaluate our method across a wide range of benchmarks for MLLMs, including high-resolution perception (V*~\cite{wu2024v}, HR-Bench 4K, and 8K~\cite{wang2025divide}), general perception (MMBench~\cite{liu2024mmbench}, VisuLogic~\cite{xu2025visulogic}, and MMVP~\cite{tong2024eyes}), and hallucination benchmarks (POPE~\cite{li2023evaluating} and Hallusionbench~\cite{guan2024hallusionbench}). A broad range of empirical results across these diverse benchmarks demonstrates that our method achieves consistent and considerable improvements over existing visual reasoning baselines. Beyond reasoning outcomes, we also assess the pixel grounding ability of SegAnswer. On segmentation benchmarks (RefCOCO~\cite{kazemzadeh2014referitgame}, RefCOCO+~\cite{kazemzadeh2014referitgame}, and RefCOCOg~\cite{mao2016generation}), our method exhibits strong pixel-level grounding performance, surpassing prior segmentation-specific approaches. In a nutshell, our contributions are threefold: 
\vspace{-5pt}
\begin{itemize}[noitemsep, topsep=0pt, partopsep=0pt, parsep=0pt]
    \item We propose to shift from rigid rectangular bounding boxes to pixel-level segmentation for localizing the region of interest more precisely.
    \item We propose the SegAnswer method, which utilizes pixel-level grounding to achieve more accurate capture of regions of interest, thereby facilitating the visual reasoning ability of MLLMs. 
    \item Our method achieves consistent improvements over previous visual reasoning methods across multiple types of evaluation benchmarks.
\end{itemize}

%% file: table_figure/fig_teaser.tex
\begin{figure*}[t]
\centering
            \begin{subfigure}[t]{0.48\textwidth}
			\centering
			\includegraphics[width=\textwidth]{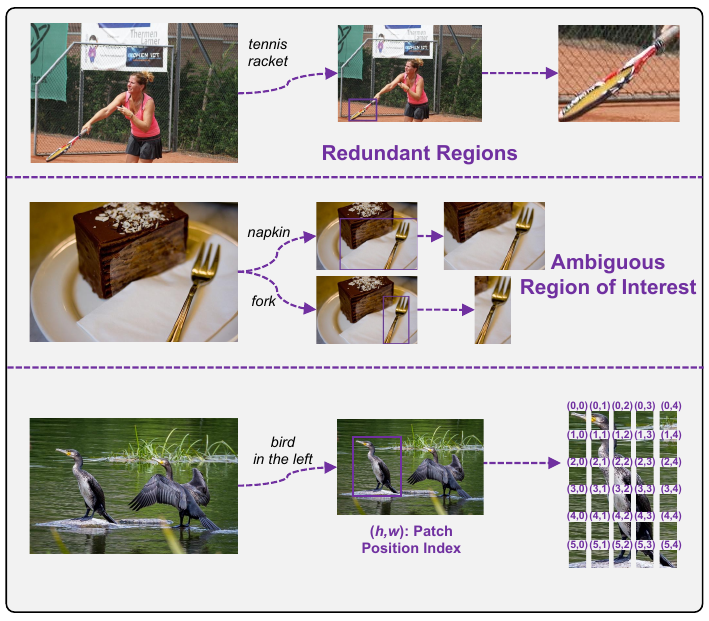}
            \caption{BBox-based zoom-in operation.}
            			\label{fig:teasera}    
	\end{subfigure}
	    \begin{subfigure}[t]{0.48\textwidth}
			\centering
			\includegraphics[width=\textwidth]{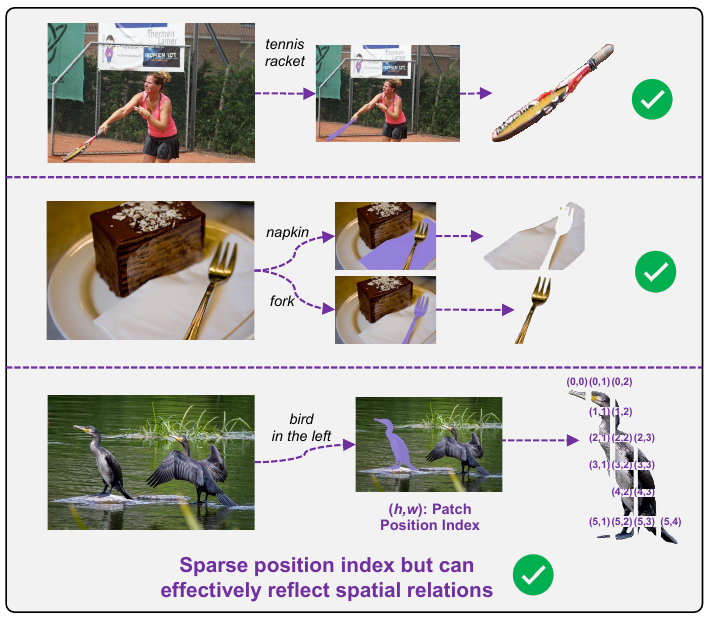}
            \caption{Our segmentation-based pixel grounding.}
            			\label{fig:teaserb}
	\end{subfigure}
    \caption{\textbf{Comparison between the BBox-based zoom-in operation and our segmentation-based pixel grounding.} \textbf{(a)} Rectangular bounding boxes inevitably introduce redundant background regions (e.g., the background around the sloping tennis racket) and fail to precisely disentangle the region of interest from overlapping objects (e.g., the fork and napkin), leading to visual noise and semantic ambiguity. \textbf{(b)} Pixel-level segmentation can precisely isolate the region of interest, effectively eliminating background noise and decoupling adjacent entities. In addition, by keeping the position index of the original image, sparse segmented image patches can also effectively reflect spatial relations.}

\end{figure*}

%% file: version3/relatedwork.tex
\section{Related Work}

\subsection{Multimodal Large Language Models}
The integration of visual encoders with powerful Large Language Models (LLMs) endowed MLLMs with visual perception capabilities~\citep{alayrac2022flamingo,awadalla2023openflamingo}. Representative architectures, such as LLaVA~\citep{liu2024visual} and BLIP-2~\citep{li2023blip}, established a prevailing paradigm where a visual connector projects image features into a text-aligned embedding space, enabling LLMs to process visual signals alongside textual inputs. While these advancements have significantly boosted performance across general vision-language tasks, the perception of visual inputs remains static. In other words, conventional MLLMs typically process visual inputs through a single forward pass as a fixed and immutable context. Consequently, the visual information available for reasoning is restricted to the initial abstraction, preventing the model from verifying details that were not prominent in the global view.

\input{table_figure/fig_method}

\subsection{MLLMs Visual Reasoning}
Beyond static perception, recent advancements in MLLMs have embraced visual reasoning, which is referred to as ``thinking with image''. These methods empower models to actively manipulate visual content by diverse operations, including drawing auxiliary lines~\citep{hu2024visual}, shifting image styles~\citep{liu2025Visual}, highlighting sub-regions~\citep{fu2025refocus}, and zooming in on specific areas~\citep{wang2025pixel,zheng2025deepeyes}. Among these operations, the zoom-in mechanism is prevalent and effective for capturing fine-grained details for the region of interest. Nevertheless, natural objects are rarely perfect rectangles, and their shapes are often irregular or non-convex. However, typical implementations of zoom-in rely on generating bounding boxes to define the region of interest. 
This BBox-based zoom-in operation would capture redundant background information and fail to effectively disentangle overlapping objects, leading to semantic ambiguity. To address these limitations, we introduce SegAnswer, a method that replaces coarse bounding boxes with precise semantic segmentation. By leveraging pixel-level grounding, SegAnswer isolates the target object accurately, eliminating visual interference and providing a clean context for subsequent reasoning.

%% file: table_figure/fig_method.tex
\begin{figure*}[t]
    \centering
    \includegraphics[width=1\linewidth]{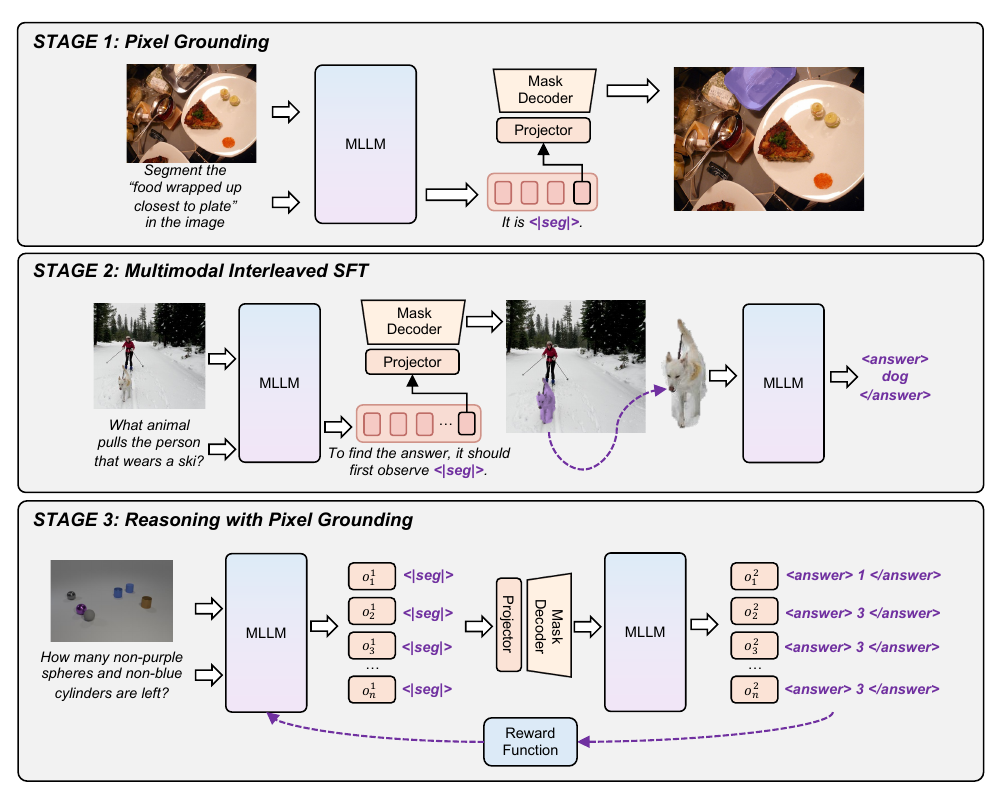}
    \caption{\textbf{Overview of our SegAnswer method.} The training pipeline progresses through three stages: \textbf{Stage 1: Pixel Grounding} aligns textual semantics with pixel-level features, training the MLLM to generate segmentation masks via a specialized \texttt{<|seg|>} token and a mask decoder. \textbf{Stage 2: Multimodal Interleaved SFT} enables the model to employ segmentation as an intermediate conversation step, using the generated mask to focus the visual context before answering. \textbf{Stage 3: Reasoning with Pixel Grounding} utilizes reinforcement learning to enhance MLLM visual reasoning by precise and finer segmented visual inputs.}
    \vspace{0.5em}
    \label{fig:method}
\end{figure*}

%% file: version3/method.tex
\section{Method}

In this paper, we introduce SegAnswer, a novel framework designed to enhance the MLLM visual reasoning by precise, pixel-level grounding. Unlike former approaches that rely on bounding boxes for visual content capture, SegAnswer empowers the MLLM to actively segment regions of interest during the reasoning process. This capability allows the model to eliminate background noise and resolve semantic ambiguities inherent in complex visual scenes. As illustrated in~\autoref{fig:method}, our training has three stages: pixel grounding, multimodal interleaved supervised fine-tuning (SFT), and reasoning with pixel grounding.

\subsection{Stage 1: Pixel Grounding}
\label{sec:stage1}

This stage aims to align textual semantics with pixel-level visual features, effectively enabling the model's pixel grounding ability, catching and segmenting the region of interest based on textual instructions.

Our base model is Qwen2.5-VL-7B~\cite{bai2025qwen2}. To enable mask prediction for segmentation, we integrate a projector and the SAM 2.1~\cite{ravi2024sam} as the mask decoder. The projector is a simple MLP, and it aims to help align the hidden tokens of the MLLM with the mask decoder. To obtain the mask based on the hidden tokens of the MLLM, a special token, \texttt{<|seg|>}, is utilized as the interface. During the forward pass, we extract the last-layer hidden states corresponding to this token. Then, these hidden states are processed by the projector and mapped into the input space of the mask decoder. Then, the mask decoder will decode the mask result as in~\cite{ravi2024sam}. An example of the prompt template is as follows:

\begin{tcolorbox}[boxrule=1pt, rounded corners, left=2mm, right=2mm, top=2mm, bottom=2mm]
\fontsize{7.5pt}{8pt}\selectfont
\textbf{Prompt 1:} \texttt{<image>}. \\
Segment the “food wrapped up closest to plate” in the image. \\
\textbf{Response 1:} It is \texttt{<|seg|>}.
\end{tcolorbox}

For this stage, the training objective is a linear combination of the next token prediction loss and segmentation-specific losses. For the segmentation-specific losses, we adopt the loss configuration from SAM 2.1~\cite{ravi2024sam}, which includes a focal loss and dice loss for mask prediction, a mean-absolute-error loss for IoU prediction, and a cross-entropy loss for object prediction. To ensure stable feature alignment, we implement a two-phase training strategy. Initially, we first freeze both the MLLM backbone and the mask decoder, updating only the newly introduced projector. Subsequently, Low-Rank Adaptation (LoRA)~\cite{hu2022lora} is applied to the MLLM. The projector, mask decoder, and LoRA layers of the MLLM are jointly fine-tuned. After Stage 1, the LoRA weights are merged into the MLLM parameters, equipping the model with the ability to perform pixel-level grounding.

\subsection{Stage 2: Multimodal Interleaved SFT}
\label{sec:stage2}
Building upon the pixel grounding ability established in Stage 1, this stage aims to evolve the model's capability to handle multimodal interleaved reasoning conversation patterns, where segmented images of pixel grounding are employed as an intermediate conversation step.

As illustrated in the prompt template below, the model is instructed to first analyze the image and question to decide if a specific region requires isolation. If affirmative, it will generate a \texttt{<|seg|>} token to execute the segmentation. Then, the segmented image will be decoded as described in~\autoref{sec:stage1}. The resulting segmented image is then fed into the model, allowing the model to observe fine-grained visual information of the region of interest.

\begin{tcolorbox}[boxrule=1pt, rounded corners, left=2mm, right=2mm, top=0mm, bottom=2mm]
\fontsize{7.5pt}{8pt}\selectfont
\textbf{Prompt 1:} \texttt{<image>}. \\
Question: What animal pulls the person that wears a ski?\\
Based on the original image and the question, reason whether there exists a region in the image that could help you answer the question better. If such a region exists, do segmentation to get the question-relevant area with the tag \texttt{<|seg|>}. \\
Then, you will receive the segmented area. Use both the segmented area and the original image to answer the question inside the \texttt{<answer>} and \texttt{</answer>} tags.\\
Format Example: \texttt{<|seg|>} OR \texttt{<answer>} final answer \texttt{</answer>}. \\
\textbf{Response 1:} To find the answer, it should first observe \texttt{<|seg|>}.\\
\textbf{Prompt 2:} \texttt{<segmented region of interest>} \\
\textbf{Response 2:} \texttt{<answer>} dog \texttt{</answer>}.
\end{tcolorbox}

It should be noted that we keep the position index of each segmented image patch as it is in the original image, as shown in~\autoref{fig:teaserb}. In this way, sparse segmented image patches can effectively reflect spatial relations. This way also aligns more seamlessly with the nature that MLLMs structure visual tokens through positional embeddings. 

In addition, the optimization of the model relies solely on the next-token prediction loss derived from the language modeling objective in this stage. We do not apply explicit supervision to the segmentation mask output. Therefore, the projector and mask decoder are frozen to preserve the segmentation ability learned in Stage 1. We unfreeze and fine-tune the full parameters of the MLLM backbone. 

By the end of this stage, the model is expected to be equipped with preliminary capabilities to handle multimodal interleaved scenarios, effectively treating segmentation as an intrinsic tool and establishing a solid behavioral initialization for subsequent reinforcement learning.

\subsection{Stage 3: Reasoning with Pixel Grounding}

This stage focuses on optimizing the reasoning strategy through Reinforcement Learning (RL) to enhance the visual reasoning ability of the MLLM. 

Unlike text-only RL, where the state consists solely of text tokens, the visual reasoning process incorporates visual tokens of segmented areas derived from pixel grounding. At each time step $t$, the $i$-th rollout sequence $o^t_i$ is defined as the interleaved sequence of text and visual history:
\begin{equation}
o^t_i = \{ \mathbf{X}_{\leq t}, \mathbf{I}_{\leq t} \} = {(X_0, I_0), \dots, (X_t, I_t) },
\end{equation}
where $\mathbf{X}_{\leq t}$ represents the accumulated text tokens, $I_{0}$ represents the visual tokens of the original image, and $\mathbf{I}_{\geq 1}$ represents the segmented images. Also, as stated in~\autoref{sec:stage2}, the position index of segmented images is also kept the same as it is in the original image, to well preserve the spatial relations. Then, given this state, the policy model (\textit{i.e.}, the MLLM) generates the next token in the sequence. This process continues iteratively until the model generates a final answer or reaches a maximum step limit.

During visual reasoning, intermediate visual operations (here is pixel grounding) lack explicit ground truth supervision. Therefore, we adopt an outcome-driven reward formulation that evaluates the entire reasoning trajectory based on result quality. The total reward has two parts: an accuracy reward $R_{\text{acc}}$, and a format reward $R_{\text{format}}$. Given a final reasoning trajectory $\tau$, the total reward is defined as:
\begin{equation}
R(\tau) = w_a * R_{\text{acc}}(\tau) + w_f* R_{\text{format}}(\tau),
\end{equation}
where $w_a$ and $w_f$ is the weight of $R_{\text{acc}}(\tau)$ and $R_{\text{format}}(\tau)$, repsectively.

For the RL algorithm, we leverage the decoupled clip and dynamic sampling Policy Optimization (DAPO)~\cite{yu2025dapo}. The effectiveness and efficiency of this algorithm have been verified across different tasks. Also, for the final multimodal trajectories, we use a token-wise masking setting that restricts loss calculation solely to model-predicted tokens, omitting the observation inputs.

After three stages, it completes the training pipeline of SegAnswer, equipping the MLLM with a visual reasoning capability with pixel grounding.

%% file: version3/experiments.tex
\section{Experiment}

\subsection{Training Data}

\textbf{Stage 1}: In this stage, we utilize multiple object segmentation datasets to enable the model with pixel grounding ability. The used datasets include RefCOCO~\cite{kazemzadeh2014referitgame}, RefCOCO+~\cite{kazemzadeh2014referitgame}, RefCOCOg~\cite{mao2016generation}, RefClef~\cite{kazemzadeh2014referitgame}, ReasonSeg~\cite{lai2024lisa}, ADE20K~\cite{zhou2017scene}, COCOStuff~\cite{caesar2018coco}, Mapillary Vistas~\cite{neuhold2017mapillary}, PACO-LVIS~\cite{ramanathan2023paco} and PASCAL-Part~\cite{chen2014detect}. \textbf{Stage 2}: In this stage, we use VisualCOT~\cite{shao2024visual} as the training data. This is a large-scale VQA dataset that contains 438k question–answer pairs, each annotated with bounding boxes that mark the critical regions needed to derive the answer. It should be noted that we do not use the bounding box supervision. We only use the ground truth answer to supervise the final result. \textbf{Stage 3}: For RL training, we utilize ViRL39K~\cite{wang2025vl} as the training data. It is a curated collection of 39k verifiable question-answer pairs for vision-language RL training. This dataset is built on top of newly collected problems and existing datasets through cleaning, reformatting, rephrasing, and verification.

\input{table_figure/table1}

\input{table_figure/table_finegrained}

\subsection{Training settings}
For Stage 1, when training the projector, the learning rate is $1e-3$. This stage uses RefCOCO, RefCOCO+, RefCOCOg, and RefClef datasets and trains the model for $5$ epochs. Then, when training LoRA layers, the projector and the mask decoder, all datasets of Stage 1 are utilized to train the model for $3$ epochs. The learning rate is $2e-5$. For Stage 2, the learning rate is $2e-6$, and the total epoch is $1$. For Stage 3, the learning rate is $2e-5$, and the total epoch is $1$. The number of rollouts is $4$. $w_a$ is $0.8$, and $w_f$ is $0.2$.

\subsection{Benchmarks}

\textbf{High-resolution perception}: We evaluate our SegAnswer method on the visual detail understanding benchmarks with high-resolution visual inputs. These benchmarks require the model have a fine-grained understanding of the high-resolution image. In this part, we adopt V*~\cite{wu2024v}, HR-Bench 4K, and HR-Bench 8K~\cite{wang2025divide} benchmarks. \textbf{General perception}: Besides the fine-grained visual understanding benchmarks, we also assess the model's ability on widely-used general perception benchmarks, including MMBench~\cite{liu2024mmbench}, VisuLogic~\cite{xu2025visulogic}, and MMVP~\cite{tong2024eyes}. \textbf{Hallucination}: We also evaluate the model's ability on typical hallucination benchmarks, including POPE~\cite{li2023evaluating} and Hallusionbench~\cite{guan2024hallusionbench}. \textbf{Segmentation benchmarks}: To verify the pixel grounding ability of our SegAnswer, we also evaluate our model on representative segmentation datasets, RefCOCO~\cite{kazemzadeh2014referitgame}, RefCOCO+~\cite{kazemzadeh2014referitgame}, and RefCOCOg~\cite{mao2016generation}.

\subsection{Main Results}

We first evaluate SegAnswer across three distinct categories of multimodal benchmarks: High-Resolution Perception, General Perception, and Hallucination Evaluation. As presented in~\autoref{tab:main}, we compare our framework against leading open-source MLLMs, including LLaVA-OneVision-9B~\cite{li2024llava} and our backbone model Qwen2.5-VL-7B~\cite{bai2025qwen2}. In addition, we also compare our method with recent MLLM visual reasoning methods, such as Pixel Reasoner~\cite{wang2025pixel} and DeepEyes~\cite{zheng2025deepeyes}, which utilize BBox-based visual perception operation.

Based on~\autoref{tab:main}, not surprisingly, the advantages of our approach are most demonstrated in high-resolution tasks where fine-grained detail is vital. Notably, on the V* benchmark, our method achieves a score of $86.4$, delivering a substantial improvement over the base Qwen2.5-VL-7B model. In addition, on benchmarks including V*, HR-Bench 4K, and HR-Bench 8K, SegAnswer achieves superior performance, surpassing the other visual reasoning baselines. These results validate that compared to rectangular bounding boxes that introduce redundant background noise, precise pixel-level segmentation effectively isolates the target, allowing the model to focus exclusively on the relevant visual features required for complex visual recognition tasks.

\input{table_figure/fig_case}

We also assess the model performance on the general perception and representative hallucination benchmarks. According to~\autoref{tab:main}, SegAnswer still demonstrates consistent improvements on these benchmarks. By reasoning with pixel grounding, the region of interest is accurately captured and segmented, and then the model can further show general perception improvement, besides the target fine-grained visual perception scenarios.

\subsection{Fine-grained Visual Perception}

To further assess the model's efficacy in handling complex visual details, more comparisons on the challenging V* benchmark are provided. V* benchmark assesses MLLMs’ ability to perform fine-grained visual detail search and relative spatial reasoning. This dataset is challenging since the visual input is high resolution, and it needs a detailed visual search in the image to answer the question correctly. 

As shown in~\autoref{tab:fine}, we further compare with several methods that target this task, including SEAL~\cite{wu2024v}, DyFo~\cite{li2025dyfo}, and Chain-of-Focus~\cite{zhang2025chain}. Based on the results, for this challenging task, our SegAnswer method exhibits considerable improvement. By employing pixel-level segmentation, SegAnswer strictly isolates the target features, thereby minimizing semantic interference and enabling more accurate visual reasoning in this fine-grained visual search scenario.

\input{table_figure/table_seg}
\input{table_figure/fig_reward}

\subsection{Qualitative Analysis}

To observe the concrete reasoning process of SegAnswer, we visualize several inference trajectories on the V* benchmark. The results are provided in~\autoref{fig:case}. These examples illustrate how the model invokes pixel grounding to observe small objects in the high-resolution image, avoiding the interference of complex backgrounds.

Based on the visualization results, our SegAnswer can not only accurately catch the single object that is related to the question (\textit{e.g.}, woman in~\autoref{fig:case2}), it can also handle the question that is related to multiple objects in the image. Consider the scenario in~\autoref{fig:case1}, where the model is asked to identify the spatial relationship between the broom and the folded chair, SegAnswer explicitly predicts the need for focus of both objects, generates the \texttt{<|seg|>} token, and produces a precise mask that isolates both the broom and the folded chair. By reasoning over this segmented region, the model correctly identifies that the broom is on the left side of the folded chair. 

In addition, according to the visualization results, compared to rigid rectangular bounding boxes, our segmentation results effectively filter out redundant background and ambiguous visual information, providing precise regions of interest. These qualitative analyses further demonstrate the effectiveness of the pixel grounding ability of SegAnswer and the reliability of the reasoning process.

\subsection{Evaluation of Pixel Grounding Capability}

Since our SegAnswer method utilizes pixel grounding to enhancing visual reasoning of MLLMs, its efficacy is intrinsically dependent on the precision of the underlying segmentation. To verify the effectiveness of our model's grounding capabilities, we evaluated SegAnswer on representative referring segmentation benchmarks: RefCOCO, RefCOCO+, and RefCOCOg. We also compare SegAnswer with other segmentation-specific methods, including LISA~\cite{lai2024lisa}, Groundhog~\cite{zhang2024groundhog}, LaSagnA~\cite{wei2024lasagna}, VideoLISA~\cite{bai2024one}, VISA~\cite{yan2024visa} and Vitron~\cite{fei2024vitron}.

Based on the results in~\autoref{tab:grounding}, despite being designed for the visual reasoning task, SegAnswer demonstrates considerable performance across different representative referring segmentation datasets. Specifically, SegAnswer achieves more substantial performance gains on the more challenging RefCOCO+ and RefCOCOg datasets compared to RefCOCO with relatively simple descriptions, demonstrating SegAnswer's effectiveness in tackling more demanding semantic comprehension scenarios. These results confirm that our training pipeline can ensure the model executes the segmentation with high quality, providing a trustworthy pixel grounding ability for complex visual reasoning.

\subsection{Learning Dynamics of RL Training}
During the RL training of Stage 3, we further observe the learning dynamics of the two reward components (the accuracy reward $R_{\text{acc}}$ and the format reward $R_{\text{format}}$) as well as the total reward. The results are shown in~\autoref{fig:reward}.

The accuracy reward curve (\autoref{fig:accr}) exhibits a consistent upward trend, rising from an initial value of approximately $0.35$ to over $0.55$. This steady ascent indicates that the model is successfully learning to leverage pixel grounding to derive correct answers. The format reward curve (\autoref{fig:formatr}) remains consistently high from the very onset of training. This stability exhibits the effectiveness of Stage 2, where the model is instructed to handle multimodal interleaved conversation scenarios, alleviating the cold-start issue before reinforcement learning begins (\textit{e.g.}, how to correctly use \texttt{<|seg|>}, \texttt{<answer>}, and \texttt{/<answer>} tags). Overall, the total reward curve also shows a consistent upward trend, indicating a steady training process.

%% file: table_figure/table1.tex
\begin{table*}[t]
\centering
\caption{\textbf{Comparison with other methods on diverse benchmarks.} We report performance across three categories: High-resolution perception (V*, HR-Bench 4K and HR-Bench 8K), General perception (MMBench, VisuLogic, MMVP), and Hallucination evaluation (POPE, HallusionBench). The base model of our SegAnswer method is Qwen2.5-VL-7B.}
\label{tab:main}
\renewcommand{\arraystretch}{1.11}
\setlength{\tabcolsep}{0.4mm}{
\begin{tabular}{c||ccc||ccc||cc}
\bottomrule
\rowcolor[HTML]{EFEFEF} 
\cellcolor[HTML]{EFEFEF}                                 & \multicolumn{3}{c||}{\cellcolor[HTML]{EFEFEF}\textbf{High-resolution perception}}           & \multicolumn{3}{c||}{\cellcolor[HTML]{EFEFEF}\textbf{General perception}}                & \multicolumn{2}{c}{\cellcolor[HTML]{EFEFEF}\textbf{Hallucination}} \\
\rowcolor[HTML]{EFEFEF} 
\multirow{-2}{*}{\cellcolor[HTML]{EFEFEF}\textbf{Model}} & V*                          & HR-4K                 & HR-8K                 & MMBench                     & VisuLogic                   & MMVP                        & POPE                             & HallusionBench                  \\ \hline \hline
LLaVA-OneVision-9B$\dag$ ~\cite{li2024llava}                                       & 71.7                        & 62.1                        & 54.5                        & 81.8                        & 22.7                        & 67.8                        & 85.1                             & 31.4                            \\
Qwen2.5-VL-7B$\dag$~\cite{bai2025qwen2}                                            & 77.5                        & 68.7                        & 63.4                        & 83.0                        & 26.1                        & 70.7                        & 86.0                             & 44.1                            \\ \hline
Pixel Reasoner$\dag$~\cite{wang2025pixel}                                           & 85.5                        & 73.9                        & 66.4                        & 84.7                        & 25.3                        & 71.1                        & 86.8                             & 44.6                            \\
DeepEyes$\dag$~\cite{zheng2025deepeyes}                                                 & 84.3                        & 73.5                        & 69.8                        & 85.4                        & 26.7                        & 71.3                        & 87.6                             & 45.3                            \\ \hline \hline
\rowcolor[HTML]{DDF5FE} 
SegAnswer                                                     & \textbf{86.4}               & \textbf{74.8}               & \textbf{71.3}               & \textbf{87.5}               & \textbf{27.1}               & \textbf{72.3}               & \textbf{87.8}                    & \textbf{46.3}                   \\
\rowcolor[HTML]{DDF5FE} 
$\Delta$ \textit{over base model}                        & {\color[HTML]{F8A102} +8.9} & {\color[HTML]{F8A102} +6.6} & {\color[HTML]{F8A102} +8.6} & {\color[HTML]{F8A102} +4.5} & {\color[HTML]{F8A102} +1.0} & {\color[HTML]{F8A102} +1.6} & {\color[HTML]{F8A102} +1.8}      & {\color[HTML]{F8A102} +2.2}     \\ \toprule
\end{tabular}}
\begin{flushleft}
    \footnotesize $\dag$ Re-evaluated using its official model and evaluation code.
\end{flushleft}
\end{table*}

%% file: table_figure/table_finegrained.tex
\begin{table}[]
\centering
\caption{\textbf{Performance comparison on the fine-grained visual perception benchmark, V*.} Other methods that target this task are further compared.}
\label{tab:fine}
\renewcommand{\arraystretch}{1.11}
\setlength{\tabcolsep}{3mm}{
\begin{tabular}{c||c}
\bottomrule
\rowcolor[HTML]{EFEFEF} 
\textbf{Model}                    & \textbf{V*}                 \\ \hline \hline
LLaVA-OneVision-9B$\dag$~\cite{li2024llava}                                       & 71.7                                \\
Qwen2.5-VL-7B$\dag$~\cite{bai2025qwen2}                                            & 77.5                                \\ \hline
SEAL~\cite{wu2024v}                                                     & 75.4                                \\
DyFo~\cite{li2025dyfo}                                                     & 81.2                                \\
Chain-of-Focus~\cite{zhang2025chain}                                           & 88.0                                \\
Pixel Reasoner$\dag$~\cite{wang2025pixel}                                           & 85.5                                \\
DeepEyes$\dag$~\cite{zheng2025deepeyes}                                                 & 84.3                                \\ \hline \hline
\rowcolor[HTML]{DDF5FE} 
SegAnswer                                                     & \textbf{86.4}                       \\
\rowcolor[HTML]{DDF5FE} 
$\Delta$ \textit{over base model}                        & {\color[HTML]{F8A102} +8.9}         \\ \toprule
\end{tabular}}
\begin{flushleft}
    \footnotesize $\dag$ Re-evaluated using its official model and evaluation code.
\end{flushleft}
\end{table}

%% file: table_figure/fig_case.tex
\begin{figure*}[t]
\centering
	    \begin{subfigure}[t]{.48\textwidth}
			\centering
			\includegraphics[width=\textwidth]{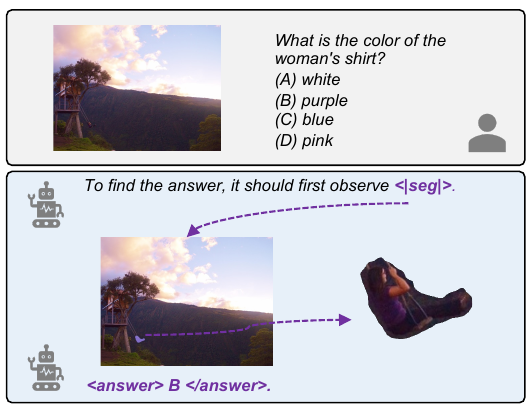}
			\caption{}
            \label{fig:case2}
	\end{subfigure}
            \begin{subfigure}[t]{.48\textwidth}
			\centering
			\includegraphics[width=\textwidth]{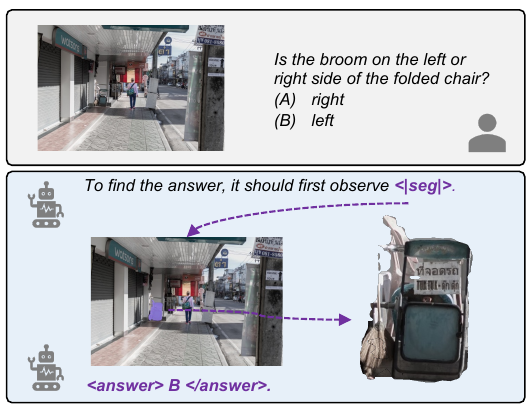}
			\caption{}
            \label{fig:case1}
	\end{subfigure}

            \begin{subfigure}[t]{.48\textwidth}
			\centering
			\includegraphics[width=\textwidth]{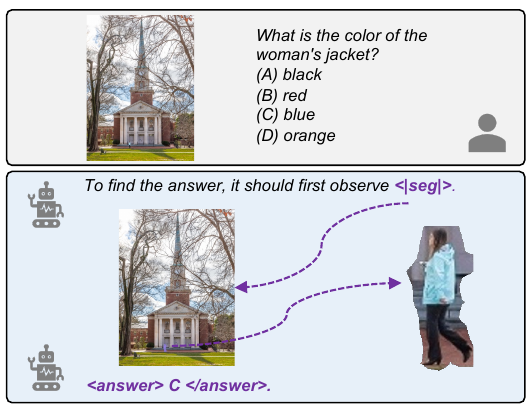}
			\caption{}
            \label{fig:case3}
	\end{subfigure}
            \begin{subfigure}[t]{.48\textwidth}
			\centering
			\includegraphics[width=\textwidth]{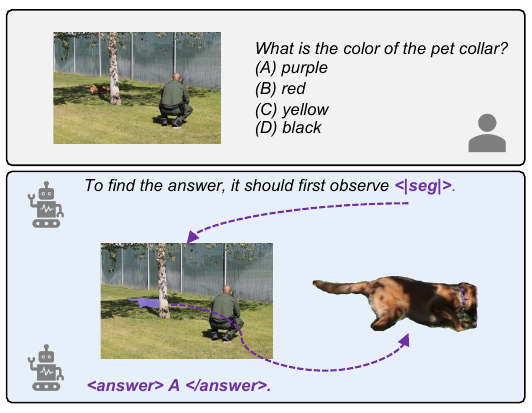}
			\caption{}
            \label{fig:case4}
	\end{subfigure}
    \vspace{-0.5em}
    \caption{\textbf{Qualitative visualization of reasoning trajectories with SegAnswer on the V* benchmark.} We showcase four examples (a-d) where the model answers fine-grained visual perception questions. In each case, the model autonomously determines the need for visual refinement, predicts the \texttt{<|seg|>} token to isolate the target object (e.g., the woman in the sample \textit{a}, and utilizes the segmented visual context to derive the correct answer.}
    \vspace{-0.5em}
    \label{fig:case}
\end{figure*}

%% file: table_figure/table_seg.tex
\begin{table*}[t]
\centering
\caption{\textbf{Performance comparison on referring segmentation benchmarks.} We evaluate the pixel grounding quality on RefCOCO, RefCOCO+, and RefCOCOg. SegAnswer is compared against other segmentation-specific methods.}
\label{tab:grounding}
\renewcommand{\arraystretch}{1.11}
\setlength{\tabcolsep}{3mm}{
\begin{tabular}{c||ccc||ccc||cc}
\bottomrule
\rowcolor[HTML]{EFEFEF} 
\cellcolor[HTML]{EFEFEF}                                 & \multicolumn{3}{c||}{\cellcolor[HTML]{EFEFEF}\textbf{RefCOCO}} & \multicolumn{3}{c||}{\cellcolor[HTML]{EFEFEF}\textbf{RefCOCO+}} & \multicolumn{2}{c}{\cellcolor[HTML]{EFEFEF}\textbf{RefCOCOg}} \\ 
\rowcolor[HTML]{EFEFEF} 
\multirow{-2}{*}{\cellcolor[HTML]{EFEFEF}\textbf{Model}} & val                 & test-A             & test-B             & val                 & test-A              & test-B             & val-u                         & test-u                        \\ \hline \hline
LISA~\cite{lai2024lisa}                                                     & 74.9                & 79.1               & 72.3               & 62.4                & 67.4                & 56.5               & 66.4                          & 68.5                          \\
Groundhog~\cite{zhang2024groundhog}                                                & 78.5                & 79.9               & 75.7               & 70.5                & 75.0                & 64.9               & 74.1                          & 74.6                          \\
LaSagnA~\cite{wei2024lasagna}                                                  & 76.8                & 78.7               & 73.8               & 66.4                & 70.6                & 60.1               & 70.6                          & 71.9                          \\
VideoLISA~\cite{bai2024one}                                                & 73.8                & 76.6               & 68.8               & 63.4                & 68.8                & 56.2               & 68.3                          & 68.8                          \\
VISA~\cite{yan2024visa}                                                     & 72.4                & 75.5               & 68.1               & 59.8                & 64.8                & 53.1               & 65.5                          & 66.4                          \\
Vitron~\cite{fei2024vitron}                                                   & 75.5                & 79.5               & 72.2               & 66.7                & 72.5                & 58.0               & 67.9                          & 68.9                          \\ \hline \hline
\rowcolor[HTML]{DDF5FE} 
SegAnswer                                                    & \textbf{79.9}       & \textbf{82.3}      & \textbf{76.9}      & \textbf{73.6}       & \textbf{78.7}       & \textbf{67.5}      & \textbf{76.0}                 & \textbf{76.4}                 \\ \toprule
\end{tabular}}
\end{table*}

%% file: table_figure/fig_reward.tex
\begin{figure*}[t]
\centering
            \begin{subfigure}[t]{.32\textwidth}
			\centering
			\includegraphics[width=\textwidth]{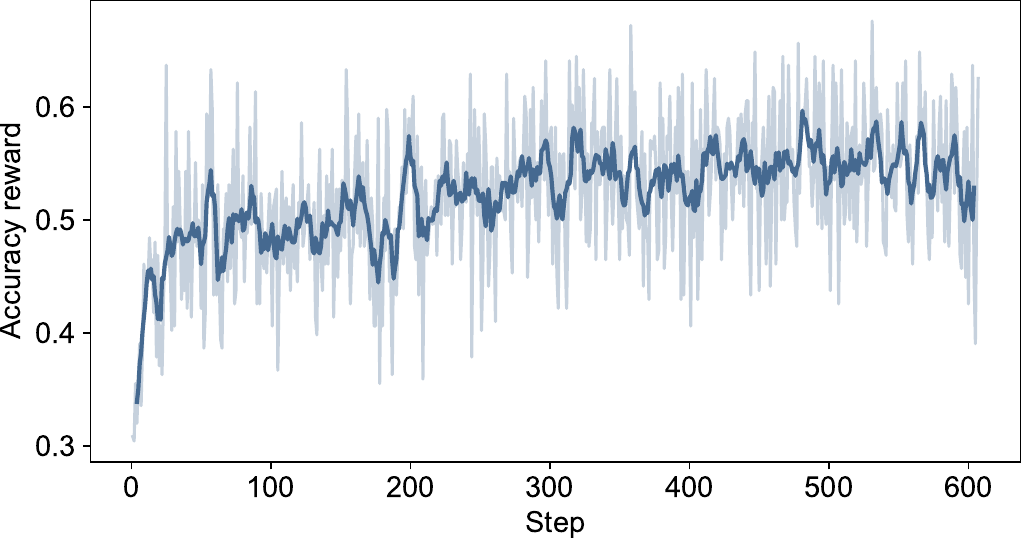}
			\caption{Accuracy reward curve.}
            \label{fig:accr}
	\end{subfigure}
	    \begin{subfigure}[t]{.32\textwidth}
			\centering
			\includegraphics[width=\textwidth]{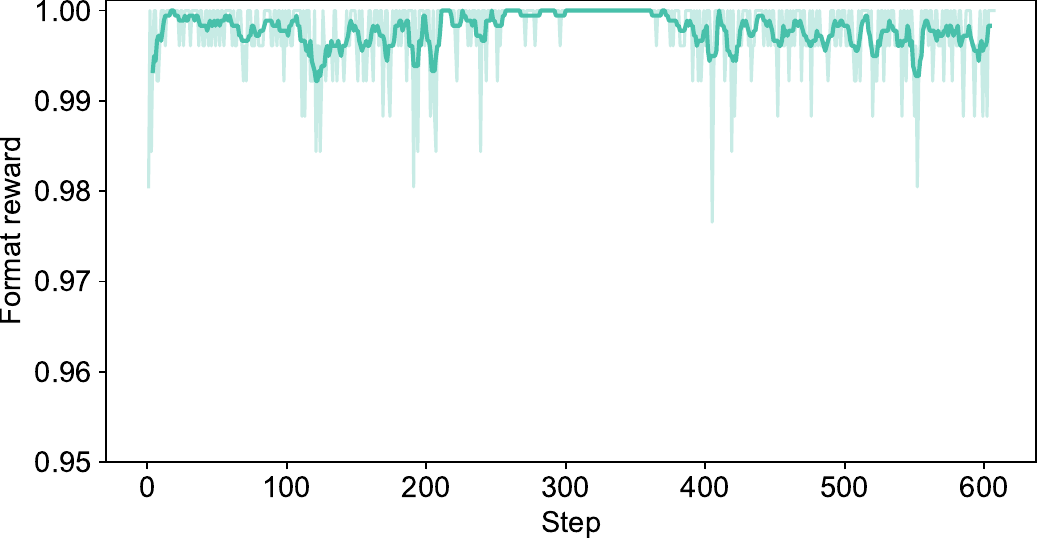}
			\caption{Format reward curve.}
            \label{fig:formatr}
	\end{subfigure}
            \begin{subfigure}[t]{.32\textwidth}
			\centering
			\includegraphics[width=\textwidth]{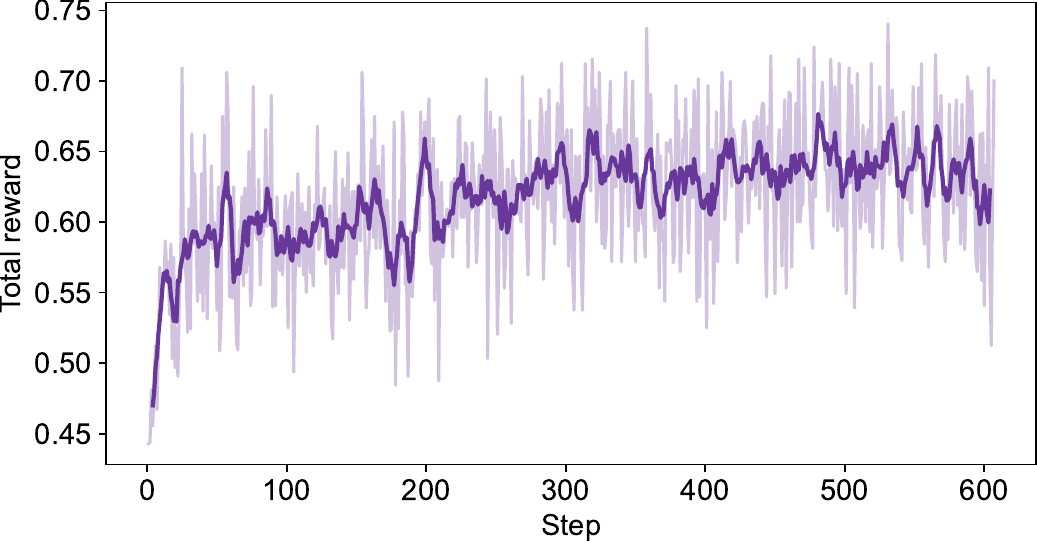}
			\caption{Total reward curve.}
            \label{fig:totalr}
	\end{subfigure}
    \caption{\textbf{Training reward curves during Stage 3: Reasoning with Pixel Grounding.} The solid lines represent the smoothed moving average, while the shaded areas indicate the raw variation at each step.}
    \label{fig:reward}
\end{figure*}

%% file: version3/conclusion.tex
\section{Conclusion}

In this work, SegAnswer is introduced to handle the fine-grained visual reasoning of MLLMs by integrating pixel-level grounding. We identify that the prevailing BBox-based zoom-in operation for visual reasoning often suffers from inherent redundancy and semantic ambiguity, particularly when processing irregular shapes or overlapping objects. By replacing inflexible bounding boxes prediction with precise semantic segmentation, SegAnswer empowers the MLLM to actively isolate and focus on exact regions of interest, thereby eliminating visual noise and resolving conflicting semantic signals. Extensive empirical evaluations across diverse benchmarks demonstrate that SegAnswer consistently shows considerable performance. The model exhibits impressive intrinsic segmentation capabilities, validating its high-quality pixel grounding ability.

%% file: example_paper.bib
@article{li2024llava,
  title={LLaVA-OneVision: Easy Visual Task Transfer},
  author={Li, Bo and Zhang, Yuanhan and Guo, Dong and Zhang, Renrui and Li, Feng and Zhang, Hao and Zhang, Kaichen and Zhang, Peiyuan and Li, Yanwei and Liu, Ziwei and others},
  journal={Transactions on Machine Learning Research},
year={2025}
}

@article{bai2025qwen2,
  title={Qwen2. 5-vl technical report},
  author={Bai, Shuai and Chen, Keqin and Liu, Xuejing and Wang, Jialin and Ge, Wenbin and Song, Sibo and Dang, Kai and Wang, Peng and Wang, Shijie and Tang, Jun and others},
  journal={arXiv preprint arXiv:2502.13923},
  year={2025}
}

@inproceedings{wu2024v,
  title={V?: Guided visual search as a core mechanism in multimodal llms},
  author={Wu, Penghao and Xie, Saining},
  booktitle={Proceedings of the IEEE/CVF Conference on Computer Vision and Pattern Recognition},
  pages={13084--13094},
  year={2024}
}

@inproceedings{li2025dyfo,
  title={Dyfo: A training-free dynamic focus visual search for enhancing lmms in fine-grained visual understanding},
  author={Li, Geng and Xu, Jinglin and Zhao, Yunzhen and Peng, Yuxin},
  booktitle={Proceedings of the Computer Vision and Pattern Recognition Conference},
  pages={9098--9108},
  year={2025}
}

@article{zhang2025chain,
  title={Chain-of-Focus: Adaptive Visual Search and Zooming for Multimodal Reasoning via RL},
  author={Zhang, Xintong and Gao, Zhi and Zhang, Bofei and Li, Pengxiang and Zhang, Xiaowen and Liu, Yang and Yuan, Tao and Wu, Yuwei and Jia, Yunde and Zhu, Song-Chun and others},
  journal={arXiv preprint arXiv:2505.15436},
  year={2025}
}

@article{wang2025pixel,
  title={Pixel reasoner: Incentivizing pixel-space reasoning with curiosity-driven reinforcement learning},
  author={Wang, Haozhe and Su, Alex and Ren, Weiming and Lin, Fangzhen and Chen, Wenhu},
  journal={arXiv preprint arXiv:2505.15966},
  year={2025}
}

@article{zheng2025deepeyes,
  title={DeepEyes: Incentivizing" Thinking with Images" via Reinforcement Learning},
  author={Zheng, Ziwei and Yang, Michael and Hong, Jack and Zhao, Chenxiao and Xu, Guohai and Yang, Le and Shen, Chao and Yu, Xing},
  journal={arXiv preprint arXiv:2505.14362},
  year={2025}
}

@inproceedings{lai2024lisa,
  title={Lisa: Reasoning segmentation via large language model},
  author={Lai, Xin and Tian, Zhuotao and Chen, Yukang and Li, Yanwei and Yuan, Yuhui and Liu, Shu and Jia, Jiaya},
  booktitle={Proceedings of the IEEE/CVF Conference on Computer Vision and Pattern Recognition},
  pages={9579--9589},
  year={2024}
}

@inproceedings{zhang2024groundhog,
  title={Groundhog: Grounding large language models to holistic segmentation},
  author={Zhang, Yichi and Ma, Ziqiao and Gao, Xiaofeng and Shakiah, Suhaila and Gao, Qiaozi and Chai, Joyce},
  booktitle={Proceedings of the IEEE/CVF conference on computer vision and pattern recognition},
  pages={14227--14238},
  year={2024}
}

@inproceedings{yan2024visa,
  title={Visa: Reasoning video object segmentation via large language models},
  author={Yan, Cilin and Wang, Haochen and Yan, Shilin and Jiang, Xiaolong and Hu, Yao and Kang, Guoliang and Xie, Weidi and Gavves, Efstratios},
  booktitle={European Conference on Computer Vision},
  pages={98--115},
  year={2024},
  organization={Springer}
}

@article{fei2024vitron,
  title={Vitron: A unified pixel-level vision llm for understanding, generating, segmenting, editing},
  author={Fei, Hao and Wu, Shengqiong and Zhang, Hanwang and Chua, Tat-Seng and Yan, Shuicheng},
  journal={Advances in neural information processing systems},
  volume={37},
  pages={57207--57239},
  year={2024}
}

@inproceedings{ravi2024sam,
  title={SAM 2: Segment Anything in Images and Videos},
  author={Ravi, Nikhila and Gabeur, Valentin and Hu, Yuan-Ting and Hu, Ronghang and Ryali, Chaitanya and Ma, Tengyu and Khedr, Haitham and R{\"a}dle, Roman and Rolland, Chloe and Gustafson, Laura and others},
  booktitle={The Thirteenth International Conference on Learning Representations},
  year={2025}
}

@inproceedings{wang2025divide,
  title={Divide, conquer and combine: A training-free framework for high-resolution image perception in multimodal large language models},
  author={Wang, Wenbin and Ding, Liang and Zeng, Minyan and Zhou, Xiabin and Shen, Li and Luo, Yong and Yu, Wei and Tao, Dacheng},
  booktitle={Proceedings of the AAAI Conference on Artificial Intelligence},
  volume={39},
  number={8},
  pages={7907--7915},
  year={2025}
}

@inproceedings{liu2024mmbench,
  title={Mmbench: Is your multi-modal model an all-around player?},
  author={Liu, Yuan and Duan, Haodong and Zhang, Yuanhan and Li, Bo and Zhang, Songyang and Zhao, Wangbo and Yuan, Yike and Wang, Jiaqi and He, Conghui and Liu, Ziwei and others},
  booktitle={European conference on computer vision},
  pages={216--233},
  year={2024},
  organization={Springer}
}

@article{xu2025visulogic,
  title={Visulogic: A benchmark for evaluating visual reasoning in multi-modal large language models},
  author={Xu, Weiye and Wang, Jiahao and Wang, Weiyun and Chen, Zhe and Zhou, Wengang and Yang, Aijun and Lu, Lewei and Li, Houqiang and Wang, Xiaohua and Zhu, Xizhou and others},
  journal={arXiv preprint arXiv:2504.15279},
  year={2025}
}

@inproceedings{tong2024eyes,
  title={Eyes wide shut? exploring the visual shortcomings of multimodal llms},
  author={Tong, Shengbang and Liu, Zhuang and Zhai, Yuexiang and Ma, Yi and LeCun, Yann and Xie, Saining},
  booktitle={Proceedings of the IEEE/CVF Conference on Computer Vision and Pattern Recognition},
  pages={9568--9578},
  year={2024}
}

@inproceedings{li2023evaluating,
  title={Evaluating Object Hallucination in Large Vision-Language Models},
  author={Li, Yifan and Du, Yifan and Zhou, Kun and Wang, Jinpeng and Zhao, Wayne Xin and Wen, Ji-Rong},
  booktitle={Proceedings of the 2023 Conference on Empirical Methods in Natural Language Processing},
  pages={292--305},
  year={2023}
}

@inproceedings{guan2024hallusionbench,
  title={Hallusionbench: an advanced diagnostic suite for entangled language hallucination and visual illusion in large vision-language models},
  author={Guan, Tianrui and Liu, Fuxiao and Wu, Xiyang and Xian, Ruiqi and Li, Zongxia and Liu, Xiaoyu and Wang, Xijun and Chen, Lichang and Huang, Furong and Yacoob, Yaser and others},
  booktitle={Proceedings of the IEEE/CVF Conference on Computer Vision and Pattern Recognition},
  pages={14375--14385},
  year={2024}
}

@inproceedings{kazemzadeh2014referitgame,
  title={Referitgame: Referring to objects in photographs of natural scenes},
  author={Kazemzadeh, Sahar and Ordonez, Vicente and Matten, Mark and Berg, Tamara},
  booktitle={Proceedings of the 2014 conference on empirical methods in natural language processing (EMNLP)},
  pages={787--798},
  year={2014}
}

@inproceedings{mao2016generation,
  title={Generation and comprehension of unambiguous object descriptions},
  author={Mao, Junhua and Huang, Jonathan and Toshev, Alexander and Camburu, Oana and Yuille, Alan L and Murphy, Kevin},
  booktitle={Proceedings of the IEEE conference on computer vision and pattern recognition},
  pages={11--20},
  year={2016}
}

@article{alayrac2022flamingo,
  title={Flamingo: a visual language model for few-shot learning},
  author={Alayrac, Jean-Baptiste and Donahue, Jeff and Luc, Pauline and Miech, Antoine and Barr, Iain and Hasson, Yana and Lenc, Karel and Mensch, Arthur and Millican, Katherine and Reynolds, Malcolm and others},
  journal={Advances in neural information processing systems},
  volume={35},
  pages={23716--23736},
  year={2022}
}

@article{awadalla2023openflamingo,
  title={Openflamingo: An open-source framework for training large autoregressive vision-language models},
  author={Awadalla, Anas and Gao, Irena and Gardner, Josh and Hessel, Jack and Hanafy, Yusuf and Zhu, Wanrong and Marathe, Kalyani and Bitton, Yonatan and Gadre, Samir and Sagawa, Shiori and others},
  journal={arXiv preprint arXiv:2308.01390},
  year={2023}
}

@article{liu2024visual,
  title={Visual instruction tuning},
  author={Liu, Haotian and Li, Chunyuan and Wu, Qingyang and Lee, Yong Jae},
  journal={Advances in neural information processing systems},
  volume={36},
  year={2024}
}

@inproceedings{li2023blip,
  title={Blip-2: Bootstrapping language-image pre-training with frozen image encoders and large language models},
  author={Li, Junnan and Li, Dongxu and Savarese, Silvio and Hoi, Steven},
  booktitle={International conference on machine learning},
  pages={19730--19742},
  year={2023},
  organization={PMLR}
}

@article{hu2024visual,
  title={Visual sketchpad: Sketching as a visual chain of thought for multimodal language models},
  author={Hu, Yushi and Shi, Weijia and Fu, Xingyu and Roth, Dan and Ostendorf, Mari and Zettlemoyer, Luke and Smith, Noah A and Krishna, Ranjay},
  journal={Advances in Neural Information Processing Systems},
  volume={37},
  pages={139348--139379},
  year={2024}
}

@article{liu2025visual,
  title={Visual Abstract Thinking Empowers Multimodal Reasoning},
  author={Liu, Dairu and Wang, Ziyue and Ruan, Minyuan and Luo, Fuwen and Chen, Chi and Li, Peng and Liu, Yang},
  journal={arXiv preprint arXiv:2505.20164},
  year={2025}
}

@article{fu2025refocus,
  title={ReFocus: Visual Editing as a Chain of Thought for Structured Image Understanding},
  author={Fu, Xingyu and Liu, Minqian and Yang, Zhengyuan and Corring, John Richard and Lu, Yijuan and Yang, Jianwei and Roth, Dan and Florencio, Dinei and Zhang, Cha},
  booktitle={ICLR 2025 Workshop on Foundation Models in the Wild},
  year={2025}
}

@article{hu2022lora,
  title={Lora: Low-rank adaptation of large language models.},
  author={Hu, Edward J and Shen, Yelong and Wallis, Phillip and Allen-Zhu, Zeyuan and Li, Yuanzhi and Wang, Shean and Wang, Lu and Chen, Weizhu and others},
  journal={ICLR},
  volume={1},
  number={2},
  pages={3},
  year={2022}
}

@article{yu2025dapo,
  title={Dapo: An open-source llm reinforcement learning system at scale},
  author={Yu, Qiying and Zhang, Zheng and Zhu, Ruofei and Yuan, Yufeng and Zuo, Xiaochen and Yue, Yu and Dai, Weinan and Fan, Tiantian and Liu, Gaohong and Liu, Lingjun and others},
  journal={arXiv preprint arXiv:2503.14476},
  year={2025}
}

@inproceedings{zhou2017scene,
  title={Scene parsing through ade20k dataset},
  author={Zhou, Bolei and Zhao, Hang and Puig, Xavier and Fidler, Sanja and Barriuso, Adela and Torralba, Antonio},
  booktitle={Proceedings of the IEEE conference on computer vision and pattern recognition},
  pages={633--641},
  year={2017}
}

@inproceedings{caesar2018coco,
  title={Coco-stuff: Thing and stuff classes in context},
  author={Caesar, Holger and Uijlings, Jasper and Ferrari, Vittorio},
  booktitle={Proceedings of the IEEE conference on computer vision and pattern recognition},
  pages={1209--1218},
  year={2018}
}

@inproceedings{neuhold2017mapillary,
  title={The mapillary vistas dataset for semantic understanding of street scenes},
  author={Neuhold, Gerhard and Ollmann, Tobias and Rota Bulo, Samuel and Kontschieder, Peter},
  booktitle={Proceedings of the IEEE international conference on computer vision},
  pages={4990--4999},
  year={2017}
}

@inproceedings{ramanathan2023paco,
  title={Paco: Parts and attributes of common objects},
  author={Ramanathan, Vignesh and Kalia, Anmol and Petrovic, Vladan and Wen, Yi and Zheng, Baixue and Guo, Baishan and Wang, Rui and Marquez, Aaron and Kovvuri, Rama and Kadian, Abhishek and others},
  booktitle={Proceedings of the IEEE/CVF Conference on Computer Vision and Pattern Recognition},
  pages={7141--7151},
  year={2023}
}

@inproceedings{chen2014detect,
  title={Detect what you can: Detecting and representing objects using holistic models and body parts},
  author={Chen, Xianjie and Mottaghi, Roozbeh and Liu, Xiaobai and Fidler, Sanja and Urtasun, Raquel and Yuille, Alan},
  booktitle={Proceedings of the IEEE conference on computer vision and pattern recognition},
  pages={1971--1978},
  year={2014}
}

@article{wang2025vl,
  title={Vl-rethinker: Incentivizing self-reflection of vision-language models with reinforcement learning},
  author={Wang, Haozhe and Qu, Chao and Huang, Zuming and Chu, Wei and Lin, Fangzhen and Chen, Wenhu},
  journal={arXiv preprint arXiv:2504.08837},
  year={2025}
}

@article{shao2024visual,
  title={Visual cot: Unleashing chain-of-thought reasoning in multi-modal language models},
  author={Shao, Hao and Qian, Shengju and Xiao, Han and Song, Guanglu and Zong, Zhuofan and Wang, Letian and Liu, Yu and Li, Hongsheng},
  journal={CoRR},
  year={2024}
}

@article{wei2024lasagna,
  title={Lasagna: Language-based segmentation assistant for complex queries},
  author={Wei, Cong and Tan, Haoxian and Zhong, Yujie and Yang, Yujiu and Ma, Lin},
  journal={arXiv preprint arXiv:2404.08506},
  year={2024}
}

@article{bai2024one,
  title={One token to seg them all: Language instructed reasoning segmentation in videos},
  author={Bai, Zechen and He, Tong and Mei, Haiyang and Wang, Pichao and Gao, Ziteng and Chen, Joya and Zhang, Zheng and Shou, Mike Zheng},
  journal={Advances in Neural Information Processing Systems},
  volume={37},
  pages={6833--6859},
  year={2024}
}
